\documentclass[10pt,twocolumn,letterpaper]{article}

\usepackage[pagenumbers]{cvpr} 
\newcommand{\qheading}[1]{\noindent\textbf{#1:}}

\usepackage{lmodern}
\usepackage{epsfig}
\usepackage{graphicx}
\usepackage{amsmath}
\usepackage{amssymb}
\usepackage{caption}
\usepackage{natbib}
\usepackage{enumitem}
\usepackage{booktabs}
\usepackage{arydshln}
\usepackage{subcaption}
\usepackage{tikz}
\usepackage{booktabs}
\usepackage{arydshln}

\definecolor{cvprblue}{rgb}{0.21,0.49,0.74}
\usepackage[pagebackref,breaklinks,colorlinks,allcolors=cvprblue]{hyperref}

\title{NIL: No-data Imitation Learning \\ by Leveraging Pre-trained Video Diffusion Models \vspace{-10pt}}

\author{%
  Mert Albaba$^{1,2}$\thanks{E-mail correspondence to: balbaba@ethz.ch} \quad Chenhao Li$^{1}$ \quad Markos Diomataris$^{1,2}$ \\ Omid Taheri $^{2}$ \quad Andreas Krause $^{1}$ \quad Michael Black $^{2}$\\ \vspace{-10pt} \\
  $^{1}$ETH Zürich \quad
  $^{2}$Max Planck Institute for Intelligent Systems
}
\begin{document}
\twocolumn[{%
  \renewcommand\twocolumn[1][]{#1}%
 \maketitle
  \vspace*{-3em}
   \begin{center}
    \centerline{ \includegraphics[trim=0mm 0mm 0mm 0mm, clip=true, width=0.95\linewidth]{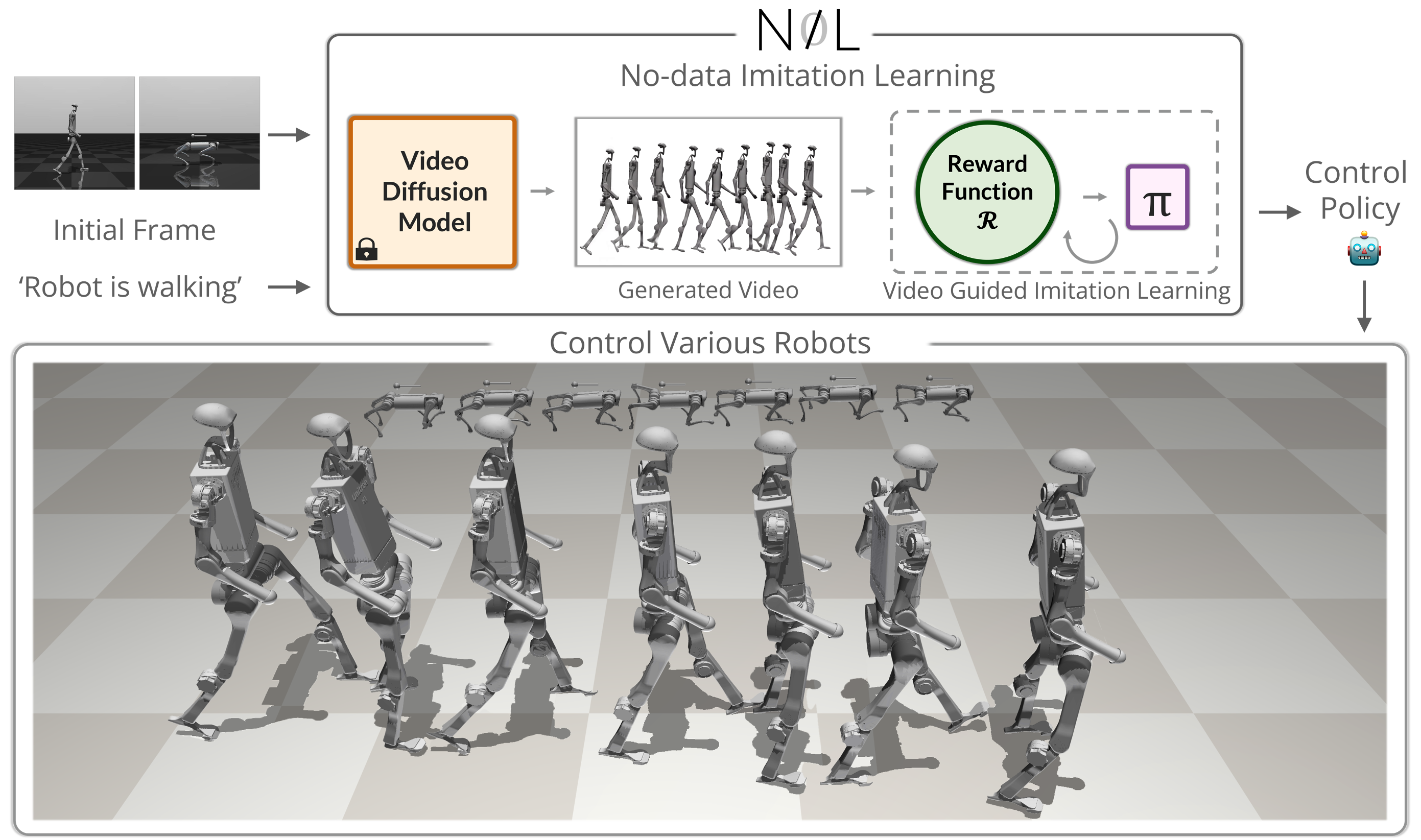}}
 \vspace*{-2.7em}
  \end{center}
  
  \begin{center}
\captionof{figure}{\textbf{NIL Overview:} First, from a single frame and a textual prompt, a pre-trained video diffusion model generates a reference video. Reinforcement learning policies are then trained to mimic the generated video and control various robots without using any external data. }
\label{fig:teaser}
\vspace*{-8pt}
\end{center}%
}]
\begin{abstract}
Acquiring physically plausible motor skills across diverse and unconventional morphologies—including humanoid robots, quadrupeds, and animals—is essential for advancing character simulation and robotics.
Traditional methods, such as reinforcement learning (RL) are task- and body-specific, require extensive reward function engineering, and do not generalize well. Imitation learning offers an alternative but relies heavily on high-quality expert demonstrations, which are difficult to obtain for non-human morphologies.
Video diffusion models, on the other hand, are capable of generating realistic videos of various morphologies, from humans to ants. 
Leveraging this capability, we propose a data-independent approach for skill acquisition that learns 3D motor skills from 2D-generated videos, with generalization capability to unconventional and non-human forms. %
Specifically, we guide the imitation learning process by leveraging vision transformers for video-based comparisons by calculating pair-wise distance between video embeddings. Along with video-encoding distance, we also use a computed similarity between segmented video frames as a guidance reward.
We validate our method on locomotion tasks involving unique body configurations. In humanoid robot locomotion tasks, we demonstrate that ``No-data Imitation Learning" (NIL) outperforms baselines trained on 3D motion-capture data.   
\looseness=-1 
\enlargethispage{1\baselineskip}
Our results highlight the potential of leveraging generative video models for physically plausible skill learning with diverse morphologies, \looseness=-1 effectively replacing data collection with data generation for imitation learning.

\end{abstract}

\section{Introduction}
\label{sec:intro}

Learning motor skills for multiple and diverse agent morphologies, including robots or animals, is essential to advance robotics and character simulation. However, enabling physically plausible skill acquisition across such a range of morphologies is a longstanding challenge.

Reinforcement learning (RL) is the most common approach for training skills in a physically plausible manner. RL trains agents within a physical simulator so that the learned behaviors inherently respect physical laws, an important property for both robotics and character simulation. However, RL requires substantial manual effort to engineer reward functions for each specific task-body pair, and poorly specified rewards lead to unintended behavior \citep{amodei2016concrete}. Imitation learning (IL) circumvents meticulous reward engineering by learning from expert demonstrations. However, IL requires high-quality 3D data, consisting of accurate joint positions and velocities. Such a high-quality 3D data is difficult to obtain for each possible morphology, particularly for non-humanoid robots and animals, where motion-capture data is scarce and expensive to collect.

This gap motivates us to explore an alternative approach that bypasses the need for high-quality demonstrations and generalizes to various morphologies. Recent advances in video diffusion models \citep{ho2022video, blattmann2023stable, zhang2023i2vgen} offer a compelling alternative. These pretrained generative models are capable of generating visually plausible video for a wide variety of tasks and morphologies. Leveraging such models could eliminate the need for curated 3D data. But while the generated videos are visually plausible, they are not always physically plausible \citep{motamed2025generative}, which hinders their usability in skill-learning. Also, learning 3D skills from generated 2D videos is challenging due to the absence of 3D information and the lack of precise action annotations.

In this work, we introduce No-data Imitation Learning (NIL): a framework that bridges this gap by integrating off-the-shelf video diffusion models with video vision transformers, and learns purely from the generated data. In other words, we propose a way to convert generated 2D videos into usable feedback for training 3D policies. Specifically, we generate reference videos using pretrained video diffusion models, and employ video vision transformers \citep{dosovitskiy2021an, arnab2021vivit} to calculate similarity between learned behaviors and generated videos. We embed both the generated video and the agent's video rendered by simulation, into the latent space of a video vision transformer. By comparing these encodings directly, we calculate a distance between videos and create a reward signal that encourages the agent to replicate key aspects of the reference motion while respecting 3D physics constraints in the simulator. Such video similarity does not provide enough guidance, since a more granular feedback is required in high-dimensional complex tasks. Therefore, we also employ image-based similarity. Specifically, we segment the agent's body in both the generated video and the simulation-rendering, creating binary masks that isolate the agent from the background. We then compute similarity scores based on the Intersection over Union (IoU) metric between the segmentation masks. These similarity scores serve as rich reward signals for IL, and along with the video-similarity, encourage the agent to produce behaviors that closely match the generated reference video, enabling it to learn skills even with unconventional body configurations.

Overall, NIL combines pretrained video diffusion models, video vision transformers, and imitation learning to enable agents to learn skills with unconventional body configurations. By directly measuring discrepancies between video encodings and segmentation masks, we provide a robust reward signal that guides the agent's learning process within a physical simulator, without relying on any curated data.
\begin{enumerate}
    \item \textbf{Autonomous Expert Data Generation:} NIL generates expert demonstrations on-the-fly using video diffusion models, conditioned on the agent's initial state and a textual task description. This approach generalizes to any task-body pair by removing any dependency on collected data.
    \item \textbf{No-data Imitation Learning:} NIL combines video vision transformers and image segmentation to create an informative reward signal from 2D videos for imitation learning. This provides a stable and effective learning guidance.
\end{enumerate} 
%
We test our approach on locomotion tasks involving diverse morphologies, including different robots (2-legged and 4-legged), for which collecting the expert data is difficult, and reward engineering is challenging. Our results show that NIL reaches the performance of motion-capture data-based imitation learning without requiring any data.

By leveraging the strengths of video diffusion models and imitation learning, our approach addresses critical bottlenecks in training agents for complex tasks, particularly when expert data is scarce or unavailable. This work opens new avenues for research at the intersection of generative modeling and reinforcement learning, with potential applications in robotics, animation, and beyond. Our code and models will be available for research purposes.

\section{Related Work}
\label{sec:rw}

\subsection{Imitation Learning}
Imitation learning (IL) has long been an attractive paradigm for training agents to mimic expert behavior. Early approaches like Behavioral Cloning \citep{bain1995framework} directly map observations to actions but are prone to compounding errors due to distributional shifts. 
Adversarial frameworks address this issue.
For example, Generative Adversarial Imitation Learning (GAIL) \citep{ho2016generative} leverages a discriminator to distinguish between expert and agent-generated trajectories, providing a robust learning signal. Subsequent methods, including InfoGAIL \citep{li2017infogail} and Adversarial Inverse Reinforcement Learning (AIRL) \citep{fu2018learning}, refine these techniques by incorporating reward inference and information-theoretic principles. Other work, such as the Variational Discriminator Bottleneck (VDB) \citep{peng2018variational} and Adversarial Motion Priors \citep{peng2021amp}, focus on stabilizing adversarial training through regularization and by incorporating pre-trained motion models. Recently, Reinforced Imitation Learning (RILe) \citep{albaba2024rile} combines inverse reinforcement learning with adversarial imitation learning to improve performance in complex settings.

While adversarial approaches are state-of-the-art in imitation learning, they have a significant drawback: the discriminator tends to overfit quickly, leading to instability during training \citep{albaba2024rile}. In contrast, our work removes this dependency by leveraging video diffusion models and direct video comparisons, providing a more stable and effective reward signal.

\subsection{Video Diffusion Models}
If we had access to limitless, high-quality, 3D training data, imitation learning would work well. Since capturing such data is challenging, is it possible to generate it?
Denoising Diffusion Probabilistic Models (DDPMs) \citep{ho2020denoising} enable high-quality image synthesis, while video diffusion models \citep{ho2022video} extend these ideas to the temporal domain, generating coherent video sequences. 
Models such as Make-A-Video \citep{singermake} and Imagen \citep{saharia2022photorealistic} demonstrate cascaded diffusion processes capture complex motion patterns from large-scale datasets. Recent developments, including Stable Video Diffusion \citep{blattmann2023stable} and I2VGen \citep{zhang2023i2vgen}, improve video quality and controllability. Additionally, recent work such as DynamiCrafter \citep{xing2024dynamicrafter} explores diffusion models tailored for dynamic scene synthesis.
Despite their impressive visual performance, these models sometimes output 2D results that are visually convincing but physically implausible \citep{motamed2025generative}, posing a challenge to using them for imitation learning.

\subsection{Video Encoders}
To exploit generated 2D data, we need to be able to use it to provide a meaningful learning signal.
Robust video encoders are critical for extracting meaningful spatio-temporal features that bridge the gap between 2D visual data and 3D behavioral understanding. Vision transformer architectures, such as ViViT \citep{arnab2021vivit} and TimeSFormer \citep{bertasius2021space}, show that transformer-based models effectively capture dynamic patterns in video data. Subsequent work, including VideoMAE-v2 \citep{wang2023videomae} and VideoMamba \citep{li2024videomamba}, advance video representation learning by employing dual masking strategies and efficient state-space models. Other approaches leverage masked autoencoding techniques \citep{feichtenhofer2022masked} to learn robust video representations. In addition, dedicated transformer architectures such as VidTr \citep{zhang2021vidtr} and DistInit \citep{girdhar2019distinit} further validate the potential of transformer-based models in video understanding.
The advances in video understanding provide robust representations that are essential for comparing generated and simulated behaviors in our framework.

\subsection{Learning from Generated Data}
There is a growing body of work that leverages generated or weakly supervised data to drive policy learning, thereby reducing the reliance on curated expert demonstrations. Early explorations in one-shot imitation learning \citep{duan2017one} and meta-imitation learning \citep{li2021meta} demonstrate that agents can learn effectively from sparse or unstructured data. More recent approaches, such as DexMV \citep{qin2022dexmv} and video language planning methods \citep{du2023video}, extend these ideas to incorporate video data directly. Complementary efforts in imitation learning from human videos \citep{xiong2021learning} and zero-shot robotic manipulation using pretrained image-editing diffusion models \citep{black2023zero} further highlight the potential of harnessing generated data. Methods like MimicPlay \citep{wang2023mimicplay} and learning universal policies via text-guided video generation \citep{du2024learning} exemplify the trend towards minimizing the dependency on meticulously collected expert data.

Notably, to the best of our knowledge, all existing approaches still rely on at least some curated data for policy training, largely because challenges remain in ensuring that visually generated data is both physically plausible and sufficiently informative for direct policy learning. NIL addresses these challenges by integrating video diffusion models within an imitation learning framework, thereby providing robust, on-the-fly expert demonstrations for agents with diverse morphologies, without the need for any curated or embodied data.


\begin{figure*}[!t]
  \centering
  \includegraphics[trim=0mm 0mm 0mm 0mm, clip=true, width=1\linewidth]{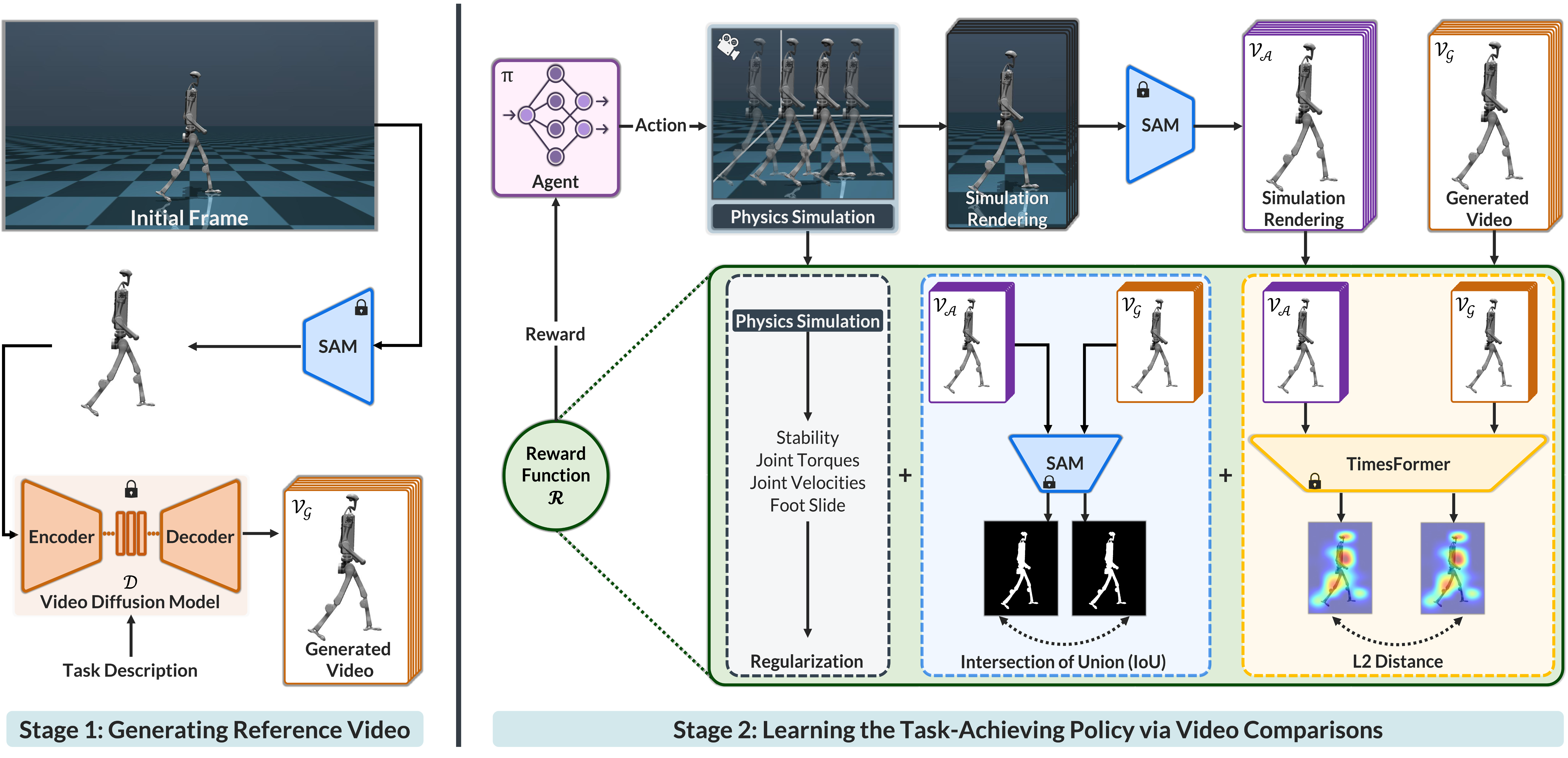}
  \caption{\textbf{NIL:} No-data Imitation Learning consists two stages. 
  \textit{Stage 1:} Render the agent’s initial frame, remove the background, and generate a reference video using a pre-trained video diffusion model conditioned on the initial frame and a textual task description.
  \textit{Stage 2:} Train a reinforcement learning agent in a physical simulation to imitate the generated video via a reward function comprising (1) video encoding similarity, (2) segmentation mask IoU, and (3) regularization for smooth behavior.
  }
  \label{fig:framework}
\end{figure*}

\section{Method}
\label{sec:method}

\subsection{Overview} 
\label{sec:overview}
No-data Imitation Learning (NIL) aims to learn physically plausible 3D motor skills from video-diffusion-model generated 2D videos.
Given a skill \( s_i \) and an embodiment \( b_j \), our goal is to learn a policy \( \pi_{s_i,b_j} \) that enables a simulated agent \( e_{b_j} \) to perform \( s_i \). Our method comprises two stages (Fig.~\ref{fig:framework}):
\begin{enumerate}
    \item \textbf{Video Generation:} A reference video \( F_{s_i,b_j} \) is generated using a pre-trained video diffusion model \( D \), conditioned on the initial 2D simulation frame \( e_0 \) and a textual prompt \( p_{s_i,b_j} \).
    \item \textbf{Policy Learning:} A reward function evaluates the similarity between the generated video \( F_{s_i,b_j} \) and a rendered simulation video \( E_{s_i,b_j} \). This reward, along with regularization for smoothness, guides the optimization of \( \pi_{s_i,b_j} \).
\end{enumerate}

\subsection{Stage 1: Video Generation} 
\label{sec:video-gen}

The video generation module uses a frozen, pre-trained video diffusion model, $D$, to generate a 2D video of the agent performing the skill $s_i$. The inputs to $D$ are: 1.~the initial frame ($e_0$) depicting the agent with embodiment $b_j$ in the simulation environment, rendered from the physical simulation at a fixed starting position, and 2.~a textual prompt ($p_{s_i,b_j}$) describing the task. The textual prompt is constructed as:
\begin{equation} 
    p_{s_i,b_j} = \text{``The } b_j \text{ agent is } s_i\text{, camera follows.''} 
\end{equation}
where $p_{b_j}$ is the name of the embodiment, e.g. Unitree H1 robot, and $p_{s_i}$ describes the skill, e.g. walking. We use a fixed camera setup; the camera follows the agent in both the generated video and the simulation rendering. The video diffusion model generates a video:
\begin{equation}
    F_{s_i,b_j} \in \mathbb{R}^{n \times H \times W \times 3} = D(p_{s_i,b_j}, e_0).
\end{equation}
The generated video frames are denoted as:
\begin{equation} 
F_{s_i,b_j} = \{ f_{0}^{(s_i,b_j)}, f_{1}^{(s_i,b_j)}, \ldots, f_{n-1}^{(s_i,b_j)}\}. 
\end{equation}
where $n$ is the number of frames, and $H$ and $W$ are the height and width of the frames, respectively.

\subsection{Stage 2: Learning the Task-Achieving Policy} 

\qheading{Video Similarity}
%
The similarity metric computes a reward signal by comparing the generated video $F_{s_i,b_j}$ with the rendered simulation video $E_{s_i,b_j} = \{e_0^{(s_i,b_j)}, e_1^{(s_i,b_j)}, \ldots, e_{k-1}^{(s_i,b_j)} \}$, where $k$ is the length of the rendered agent video. The objective is to extract meaningful learning signals from the 2D-generated video to guide the acquisition of 3D motor skills. The computation involves three steps: 1.~segmentation and masking; 2.~video encoding; and 3.~similarity computation.




\qheading{Segmentation and Masking} We segment the agent from both $F_{s_i,b_j}$ and $E_{s_i,b_j}$. For \(F_{s_i,b_j}\), we use the Segment Anything Model 2 (SAM) to obtain masks as $M^F = \{M^F_0, M^F_1, \dots, M^F_{n-1}\}$,
and for \(E_{s_i,b_j}\), segmentation masks \(M^E = \{M^E_0, M^E_1, \dots, M^E_{k-1}\}\) are provided by the simulator. Masked frames are denoted as $f^{M,(s_i,b_j)}_t$, $e^{M,(s_i,b_j)}_t$. The segmented videos are thus represented as \(F_{s_i,b_j}^M\) and \(E_{s_i,b_j}^M\).



\vspace{2pt}


\qheading{Video Encoding} To capture spatiotemporal dynamics for both the generated and rendered videos, we employ a pre-trained TimeSformer encoder \(T\) (trained on Kinetics-400). For each time step \(t\), we extract an 8-frame clip from the masked video \(F_{s_i,b_j}^M\) as:
\[
C^{(s_i,b_j)}_t =
\begin{cases}
\{ f^{M,(s_i,b_j)}_0, \ldots, f^{M,(s_i,b_j)}_0, f^{M,(s_i,b_j)}_t \}, & t < 7, \\
\{ f^{M,(s_i,b_j)}_{t-7}, \ldots, f^{M,(s_i,b_j)}_t \}, & t \geq 7.
\end{cases}
\]
Each clip is passed through \(T\) to obtain the embedding:
\[
z^{F,(s_i,b_j)}_t = T\Big( C^{(s_i,b_j)}_t \Big).
\]
An analogous procedure yields \(z^{E,(s_i,b_j)}_t\) for \(E_{s_i,b_j}^M\).


Each 8-frame clip is then passed through the TimeSformer encoder:
\[
z^{F,(s_i,b_j)}_t = T\Big( C^{(s_i,b_j)}_t \Big), \quad z^{E,(s_i,b_j)}_t = T\Big( \tilde{C}^{(s_i,b_j)}_t \Big),
\]
where $z^{F,(s_i,b_j)}_t,\, z^{E,(s_i,b_j)}_t \in \mathbb{R}^{d}$ are the embeddings derived from the last hidden states of the transformer. The encoder first divides each frame into non-overlapping patches that are linearly embedded and enriched with positional encodings. Subsequent transformer blocks then apply multi-head self-attention across both spatial and temporal dimensions, allowing the network to capture dynamic interactions among patches. This hierarchical attention mechanism yields a comprehensive representation of the input clip.


\subsubsection{Reward Function}
\label{sec:similarity}

The reward score at each frame $t$ is computed by combining video similarity, image-based similarity, and regularization.
\qheading{a) Video Similarity} The video similarity at time step $t$ is defined as the negative Euclidean (L2) distance between the corresponding embeddings of the generated and rendered videos:
\[
S_{v,t} = -\Big\| z^{F,(s_i,b_j)}_t - z^{E,(s_i,b_j)}_t \Big\|_2.
\]
%

\qheading{b) Image-Based Similarity} We compute the Intersection over Union (IoU) between the segmentation masks of the generated and rendered videos:
\begin{equation}
  S_{M_t} = \frac{\sum_{k,l} M^F_t(k,l) \cdot M^E_t(k,l)}{\sum_{k,l} M^F_t(k,l) + M^E_t(k,l) - M^F_t(k,l) \cdot M^E_t(k,l)}
\end{equation}
where $k = 1, 2, \ldots H$, and $l = 1, 2, \ldots W$ denote pixel locations vertically and horizontally, respectively. $M^F_t{(k,l)} = 1$ if the pixel at the coordinate $k,l$ stays inside the agent body mask, and $0$ otherwise. The IoU score ranges between 0 and 1, with higher values indicating greater similarity between the masks.

\qheading{c) Regularization} To ensure smooth behavior, we introduce an aggregated regularization term, $\mathcal{P}_t$. These regularization components are standard in robotic control frameworks and ensure that the learned policy adheres to physically realistic constraints. Specifically, we define:
\[
\mathcal{P}_t = P_{J,t} + P_{A,t} +  P_{V,t} +   P_{F,t} +  P_{S,t},
\]
where $P_{J,t}$ penalizes the sum of joint torques to discourage aggressive actuation, $P_{A,t}$ penalizes large differences between consecutive actions, $P_{V,t}$ discourages high angular joint velocities, $P_{F,t}$ penalizes scenarios where a foot is in contact with the ground and moving while offering a slight reward when a foot remains off the ground for an extended duration, and $P_{S,t}$ enforces stability by penalizing excessive tilting of the torso.

\qheading{d) Combined Reward} The overall reward at each time step $t$ is computed as a weighted sum of the video similarity, the image-based similarity, and the aggregated penalty:
\[
R_t = \alpha\, S_{v,t} + \beta\, S_{M,t} + \gamma\, \mathcal{P}_t,
\]
where $\alpha$, $\beta$, and $\gamma$ are scalar weights that balance the contributions of each term. This composite reward effectively aligns the rendered simulation with the generated video while promoting smooth and physically plausible agent behavior.

\subsection{Policy Learning} 
\label{sec:policy}

The end goal of NIL is to learn a policy, $\pi_{s_i,b_j}$ that maximizes the expected cumulative imitation reward derived from the similarity scores defined in Sec.~\ref{sec:similarity}. In contrast to state-of-the-art imitation learning approaches combining discriminators with reinforcement learning, we directly maximize the imitation reward using entropy-regularized off-policy actor-critic reinforcement learning. This change eliminates the need for adversarial training and simplifies the learning process.

At each time step $t$, the agent receives an observation $o_t \in \mathcal{O}$, which comprises joint positions and velocitie, and selects an action $a_t \in \mathcal{A}$ (i.e., the torques to be applied to the joints) according to the policy $\pi_{s_i,b_j}(a_t|o_t)$. The environment then provides an imitation reward, defined as:
\[
R_t = \alpha\, S_{v,t} + \beta\, S_{M,t} + \gamma\, \mathcal{P}_t,
\]
where $S_{v,t}$ is the video similarity, $S_{M,t}$ is the image-based similarity, $\mathcal{P}_t$ is the aggregated penalty (see Sec.~3.3.1.3), and $\alpha$, $\beta$, and $\gamma$ are scalar weights.

The overall objective is to maximize the expected cumulative discounted reward:
\[
J(\pi_{s_i,b_j}) = \mathbb{E}_{\pi_{s_i,b_j}}\left[\sum_{t=0}^{\infty} \gamma^t R_t\right],
\]
with $\gamma \in [0,1)$ representing the discount factor.

In the entropy-regularized actor-critic framework, the policy is optimized by maximizing a soft value function that includes an entropy term to encourage exploration:
\[
\max_{\pi} \, \mathbb{E}_{(o,a)\sim\pi}\left[\sum_{t=0}^{\infty} \gamma^t \left( R_t + \alpha_{\text{ent}}\, \mathcal{H}\big(\pi(\cdot|o_t)\big) \right) \right],
\]
where $\mathcal{H}\big(\pi(\cdot|o_t)\big)$ denotes the entropy of the policy at state $o_t$, and $\alpha_{\text{ent}}$ is a temperature parameter controlling the trade-off between reward maximization and exploration.

Both the actor (policy network) and the critics (Q-value networks) are updated concurrently using off-policy data stored in a replay buffer. This setup allows the agent to effectively leverage the dense imitation reward signal—derived from the similarity metrics—to reproduce expert motion patterns as observed in the reference video $F_{s_i,b_j}$. Importantly, while we employ BRO \citep{nauman2025bigger} in our experiments, this entropy-regularized actor-critic formulation is generic and can be instantiated with any suitable algorithm.

\qheading{Temporal Alignment} The action frequency resolution is different in the generated video when compared with the video rendered using the physical simulation since the simulator-rendered videos are very high resolution in terms of action frequency to allow fine-grained controls. Therefore, during the policy learning, to align the temporal dynamics between the simulation and the generated video, we define a mapping between simulation timesteps and video frames. Let $k$ be the number of simulation timesteps corresponding to one video frame. Then, for every $k$ simulation steps, we compute the reward $S_t$ and assign it uniformly to those steps.

\subsection{NIL}
No-data Video Imitation (NIL) provides a method for 3D motor skill acquisition from generated 2D videos by integrating video diffusion models and a discriminator-free imitation learning framework. By leveraging a pretrained video diffusion model, we generate expert demonstrations in the form of 2D videos on-the-fly. These videos offer a visual reference for the desired behavior without any need for manually collected expert data.

To extract meaningful learning signals from the generated videos, we employ a similarity metric that combines video-encoding distance and image-based segmentation similarity between the generated video and the rendered simulation video. This metric serves as a reward function guiding the agent's learning process. NIL replaces the adversarial discriminator used in state-of-the-art imitation learning methods with this similarity metric, and simplifies the training process while enhancing the stability.

Through imitation learning, the agent optimizes its policy to maximize the similarity score, effectively learning the desired skill by imitating the motion patterns depicted in the generated video. NIL's general framework allows it to be applied across diverse and unconventional morphologies, enabling agents to acquire complex skills without reliance on pre-collected expert demonstrations. NIL offers extensive generalization capability to a wide range of tasks and embodiments.
\section{Experiments}
\label{sec:experiments}
In this section, we evaluate the performance of No-data Imitation Learning (NIL) for 3D motor skill acquisition using autonomously generated 2D videos. We perform three ablation studies to analyze NIL:

\begin{itemize}
    \item \textbf{Reward Component Ablation:} We analyze the impact of individual reward components on the learning performance.
    \item \textbf{Diffusion Model Comparison:} We compare several pretrained video diffusion models to determine which one provides the most effective reference demonstrations for imitation learning.
    \item \textbf{Improving Diffusion Models:} We assess how incremental advancements in video diffusion models affect the quality of the learned behaviors.
\end{itemize}
\noindent Then, we evaluate the performance of NIL in challenging robotic control tasks: 
\begin{itemize}
    \item \textbf{Continuous Control of Various Robots:} Learning to walk with five different robot embodiments, each of which has different unique configurations and challenges.
\end{itemize}

\begin{figure*}[!t]
  \centering
  \raisebox{-8.1mm}{%
  \begin{minipage}{0.32\textwidth}
    \centering
    \includegraphics[width=\linewidth, trim={0 0 0 0}, clip]{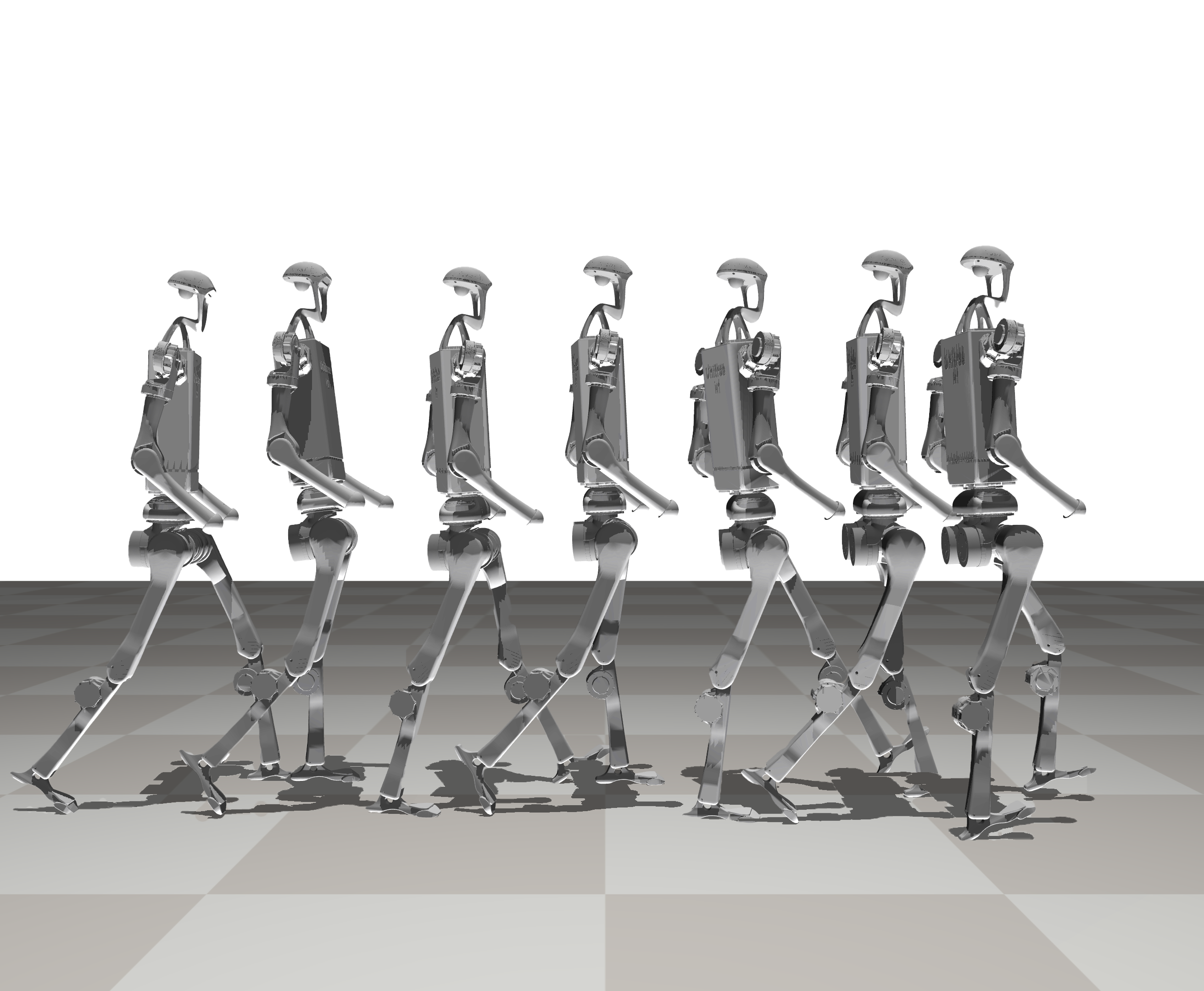}
    \subcaption{All components}
    \label{fig:allcomponents}
  \end{minipage}%
  }
  \hfill
  \begin{minipage}[t]{0.67\textwidth}
    \centering
    \begin{minipage}[t]{0.33\textwidth}
      \centering
      \includegraphics[width=\linewidth]{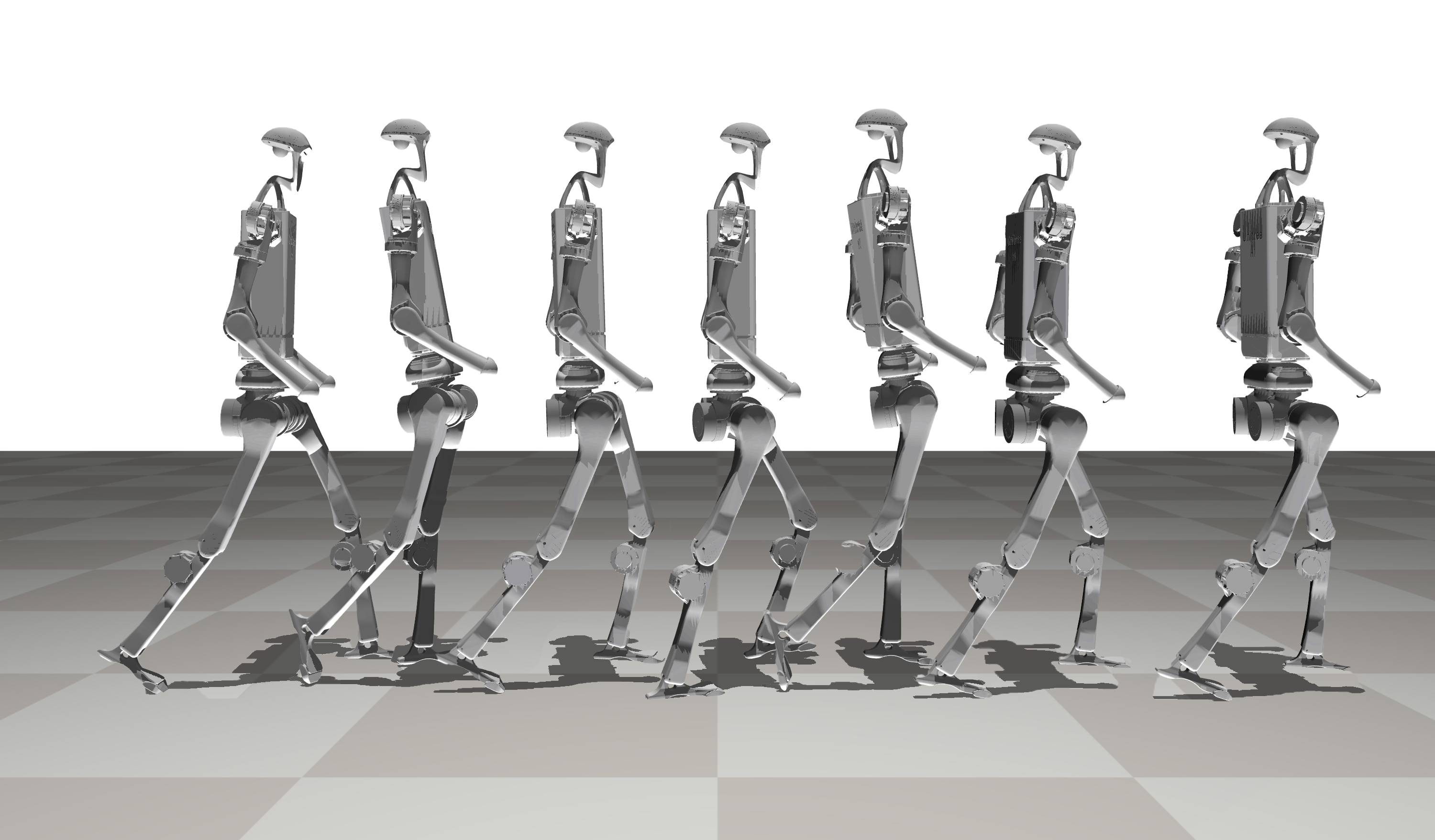}
      \subcaption{Without regularization}
      \label{fig:noreg}
    \end{minipage}%
    \hfill
    \begin{minipage}[t]{0.33\textwidth}
      \centering
      \includegraphics[width=\linewidth]{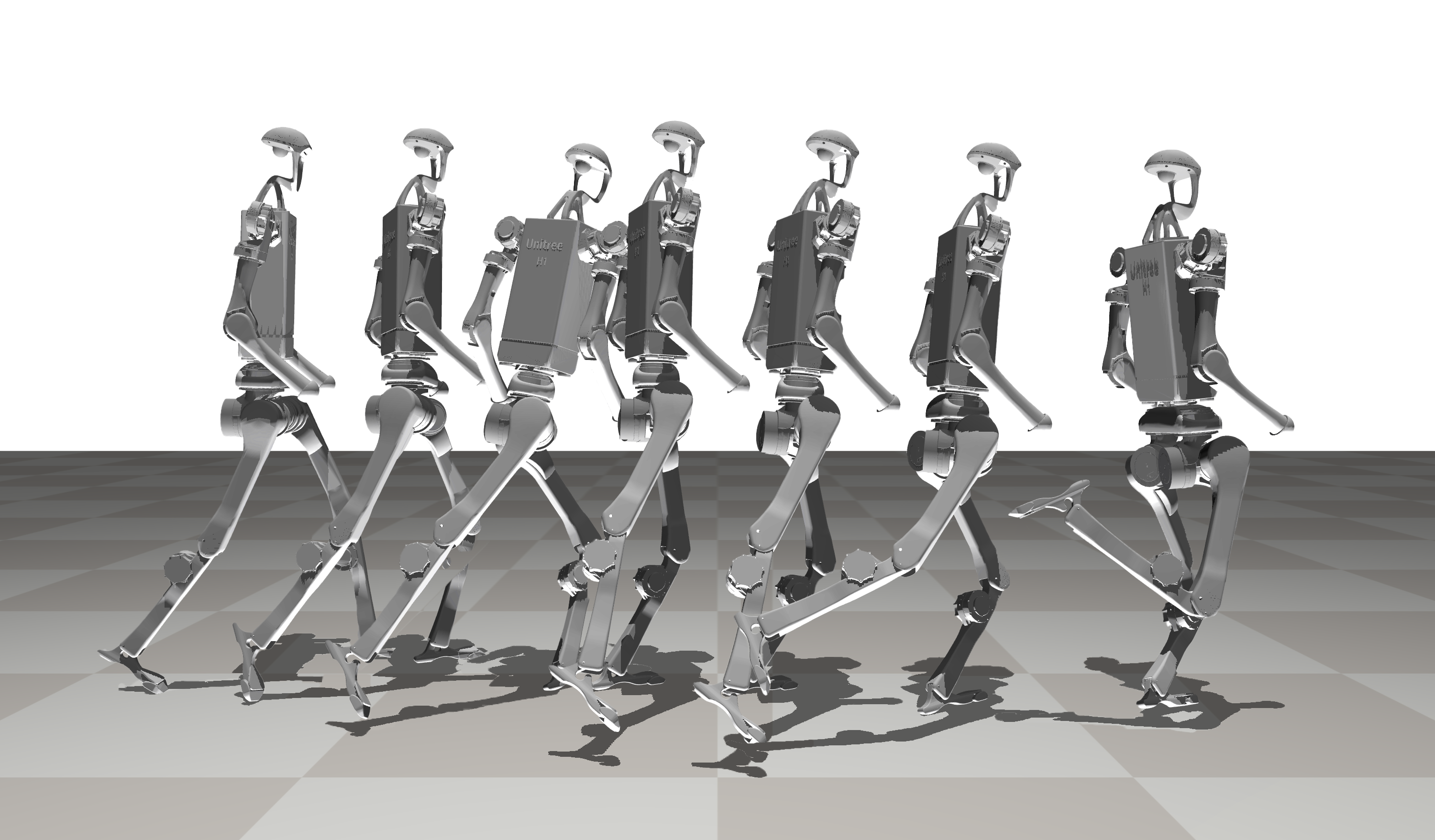}
      \subcaption{Without IoU}
      \label{fig:noiou}
    \end{minipage}%
    \hfill
    \begin{minipage}[t]{0.33\textwidth}
      \centering
      \includegraphics[width=\linewidth]{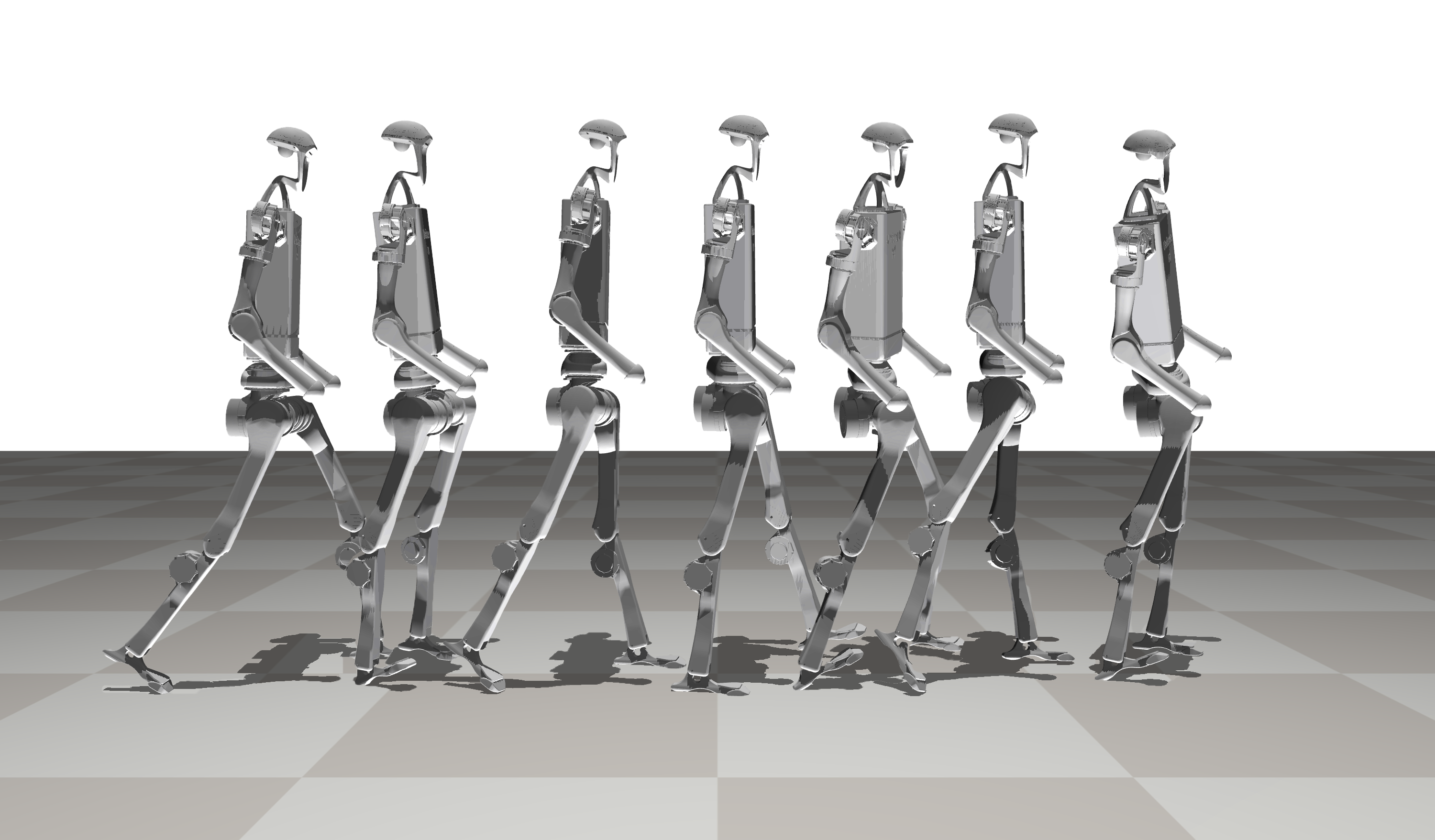}
      \subcaption{Without video similarity}
      \label{fig:novideo}
    \end{minipage}
    
    \vspace{0.3em} 
    
    \begin{minipage}[t]{0.33\textwidth}
      \centering
      \includegraphics[width=\linewidth]{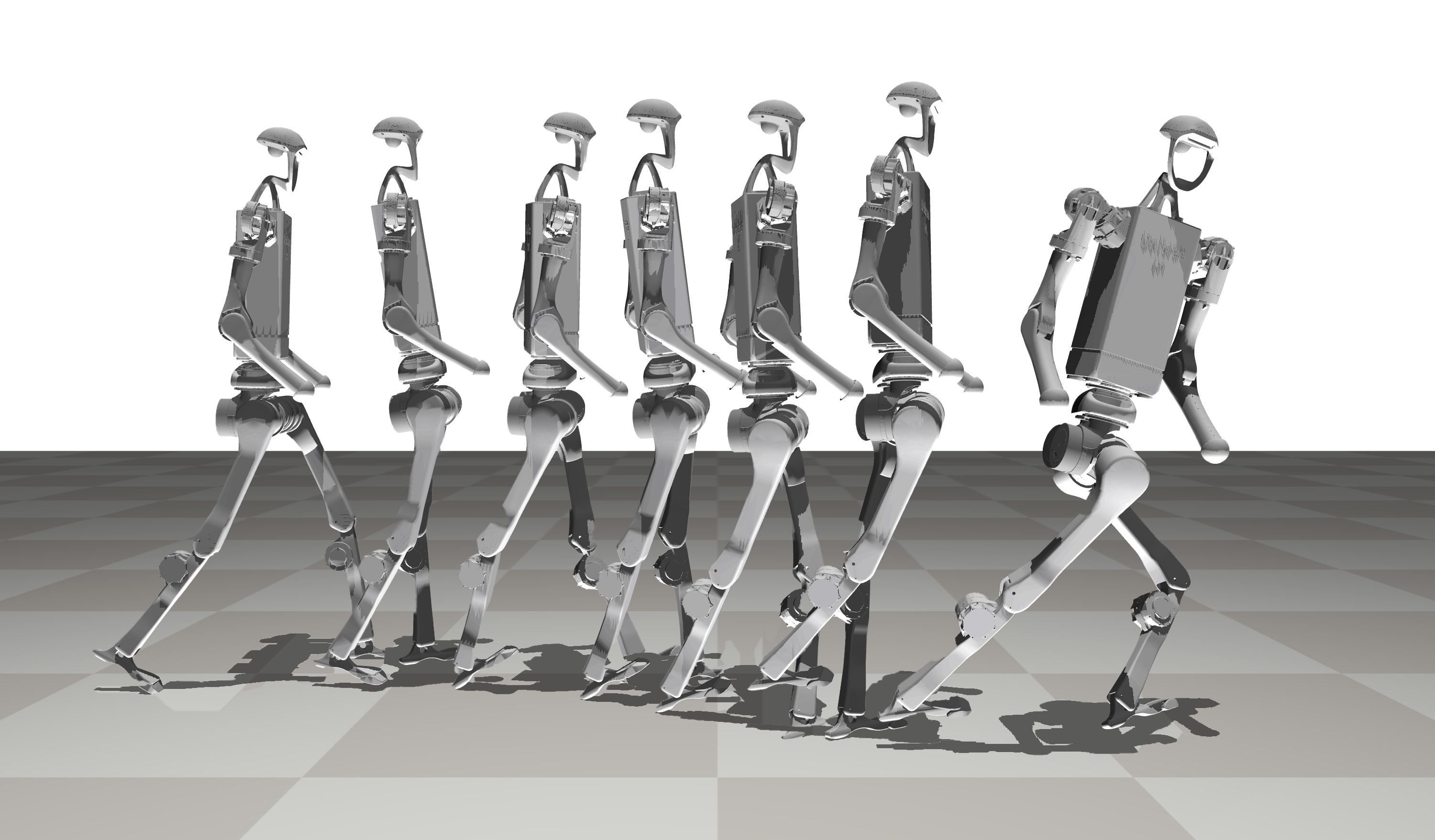}
      \subcaption{Only regularization}
      \label{fig:onlyreg}
    \end{minipage}%
    \hfill
    \begin{minipage}[t]{0.33\textwidth}
      \centering
      \includegraphics[width=\linewidth]{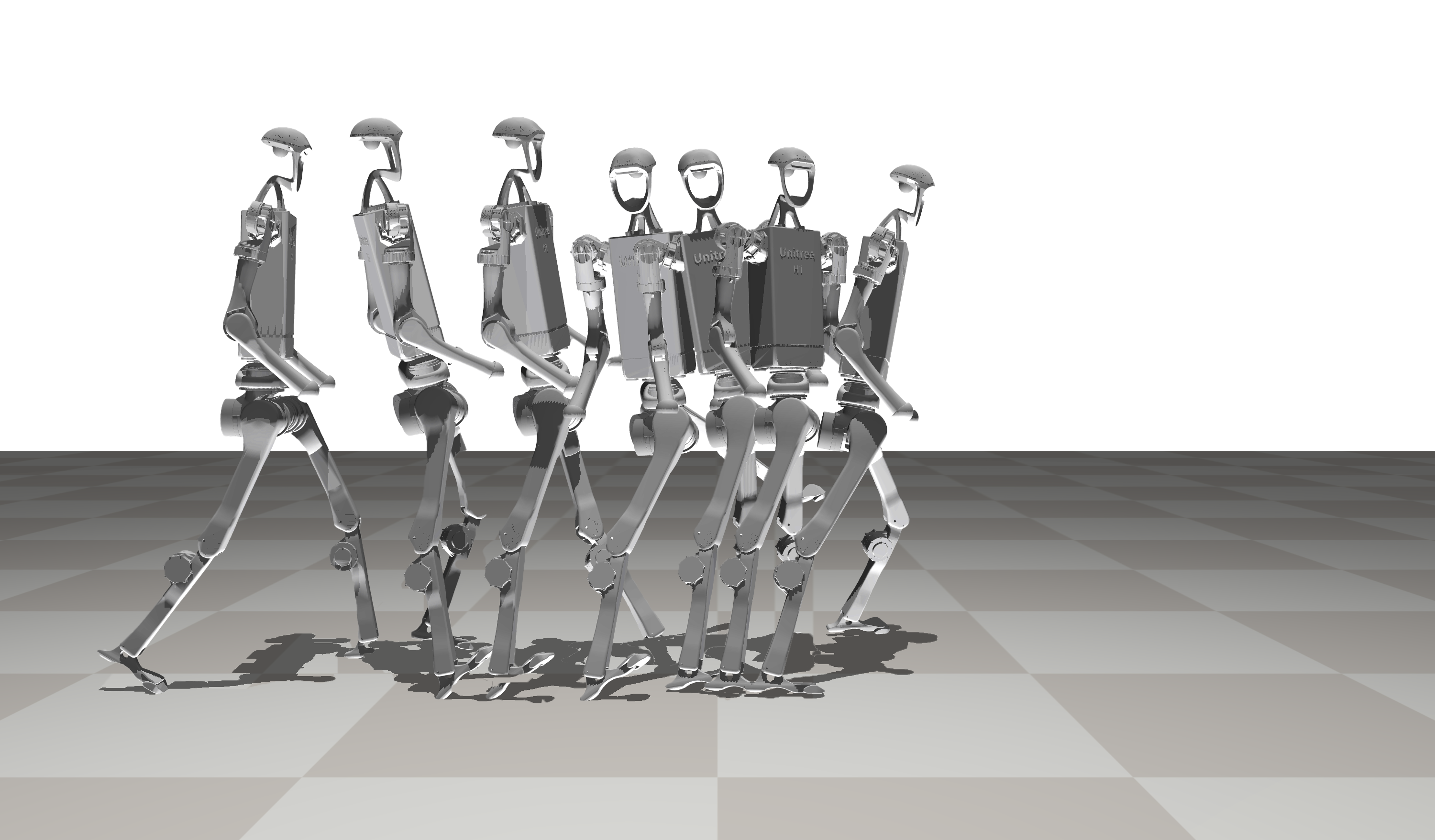}
      \subcaption{Only IoU}
      \label{fig:onlyiou}
    \end{minipage}%
    \hfill
    \begin{minipage}[t]{0.33\textwidth}
      \centering
      \includegraphics[width=\linewidth]{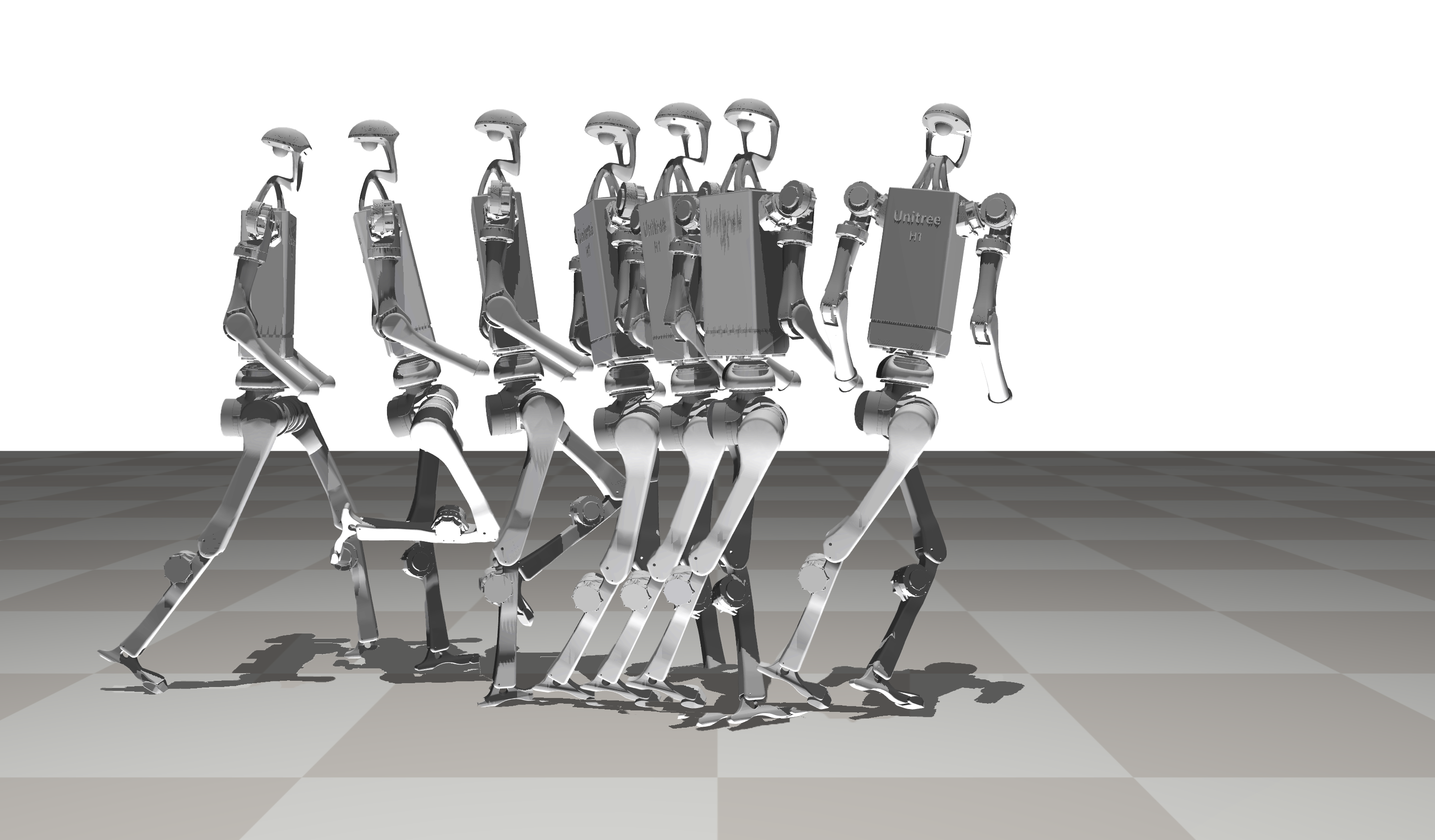}
      \subcaption{Only video similarity}
      \label{fig:onlyvideo}
    \end{minipage}
  \end{minipage}
  
  \caption{\textbf{Reward Components:} Ablation of the reward function. \textbf{(a) All components:} All components are employed, and agent learns to walk well. \textbf{(b) Without regularization:} The resulting motion is jittery. \textbf{(c) Without IoU:} The learned behavior is distorted slightly. \textbf{(d) Without video similarity:} The walking is slower, and jittery. \textbf{(e) Only regularization:} Agent fails to walk straight, and employs suboptimal large leg movements. \textbf{(f) Only IoU:} Agent fails to walk forward continuously. \textbf{(g) Only video similarity:} Agent walks in a jittery way, and stops midway while walking.}
  \label{fig:reward_ablation}
  
\end{figure*}

\noindent \textbf{Baselines:}  
Since NIL is the only method that relies solely on generated data (without any curated or same-embodiment collected data), we compare NIL against both upper and lower baselines. As upper baselines, we employ three state-of-the-art imitation learning methods: AMP \cite{peng2021amp}, GAIfO \cite{torabi2018generative}, and DRAIL \cite{lai2025diffusion}. As a lower baseline, we consider Behavioral Cloning (BC \cite{bain1995framework}) and Behavioral Cloning from Observations (BCO \cite{torabi2018behavioral}). All baselines are trained using motion-capture data from \cite{al2023locomujoco} that is adapted to the simulation domain, with perfect joint correspondence. These perfect joint correspondences are used to calculate a reward for the learning agent, and combined with regularization parameters. For AMP, we define velocity tracking reward, which rewards the agent for reaching the velocity of 1m/s.  We provide details regarding quantification metrics in Supplementary Materials.

\subsection{Reward Components}

To understand the contribution of each reward term, we train NIL on a walking task on the UnitreeH1 humanoid robot. We evaluate performance using two metrics: (a) the environment reward: evalautes the speed and stability of the policy and (b) the motion similarity score: quantifies how closely the learned motion matches motion-capture data.  

\begin{table}[!h]
\vspace{-0pt}
\centering
\caption{\textbf{Reward Ablation}: We analyze effects of each reward function component on the performance of NIL.}
\vspace{-3pt}
\label{table:reward_ablation}
\begin{tabular}{ccc} 
\toprule
                                                              & Env.~Reward $\uparrow$~ & MoCap Loss~$\downarrow$  \\ 
\hline
\begin{tabular}[c]{@{}c@{}}NIL\\(all components)\end{tabular} & 396.1                   & 46.4                     \\ 
\hline
w/o Reg.                                                  & 382.4                   & 44.9                     \\
w/o IoU Score                                             & 381.4                   & 82.9                     \\
w/o Video Sim.                                            & 387.3                   & 54.3                     \\ 
\hdashline[1pt/1pt]
only Reg.                                                     & 363.6                   & 93.5                     \\
only IoU Score                                                & 328.4                   & 101.6                    \\
only Video Sim.                                               & 369.6                   & 56.8                     \\ 
\hline
Expert                                                        & 400                     & 0                        \\
\bottomrule
\end{tabular}
\vspace{-8pt}
\end{table}

Table \ref{table:reward_ablation} presents quantitative results, Figure \ref{fig:reward_ablation} shows qualitative demonstrations. First, we analyze how the lack of individual components affects NIL. Overall, regularization helps NIL to smoothen the learned motions, while both image-based and video-based similarity scores helps the agent to understand essentials of walking. 

Second, we evaluate whether isolated components of the reward function enable imitating motions in generated reference videos. With only video similarity, NIL achieves a reasonable performance, albeit fails to generate visually plausible motions. In contrast, using only regularization or IoU rewards results in poor-performing policies.

\subsection{Diffusion Models for Imitation Learning}
We evaluate the impact of different video diffusion models on imitation learning. To our knowledge, this is the first study to compare various diffusion models for usability in imitation learning. We consider five open- and closed-source video diffusion models: Kling AI, Pika, Runway Gen-3, OpenAI Sora, and Stable Video Diffusion (SVD) \citep{blattmann2023stable}. For each model, we generate reference videos for the UnitreeH1 walking task.

\begin{figure*}[!ht]
    \centering
    \begin{subfigure}{0.495\linewidth}
        \centering
        \includegraphics[width=\linewidth]{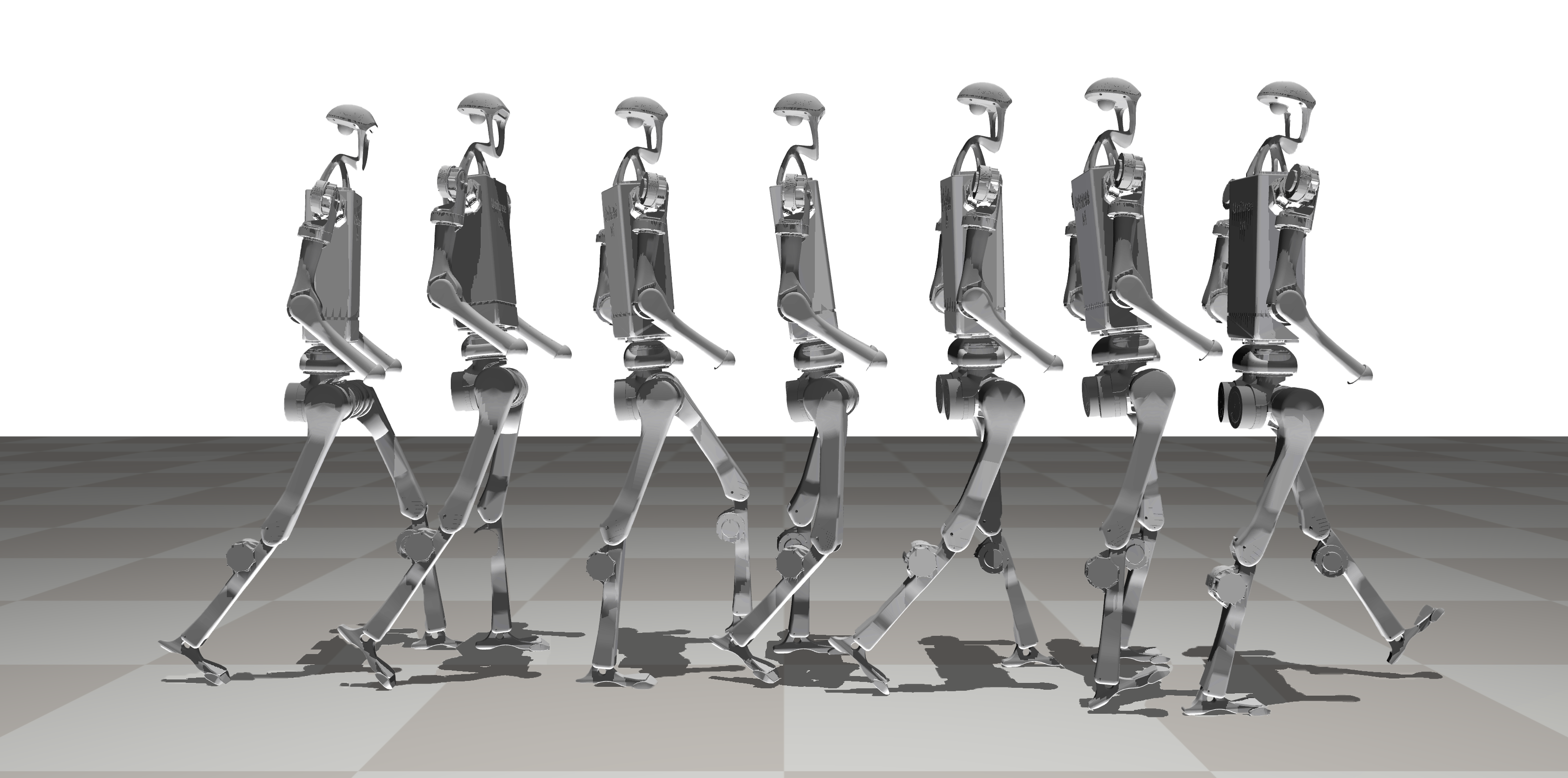}
        \caption{NIL trained on Kling v1.6}
        \label{fig:kling16}
    \end{subfigure}
    \hfill
    \begin{subfigure}{0.495\linewidth}
        \centering
        \includegraphics[width=\linewidth]{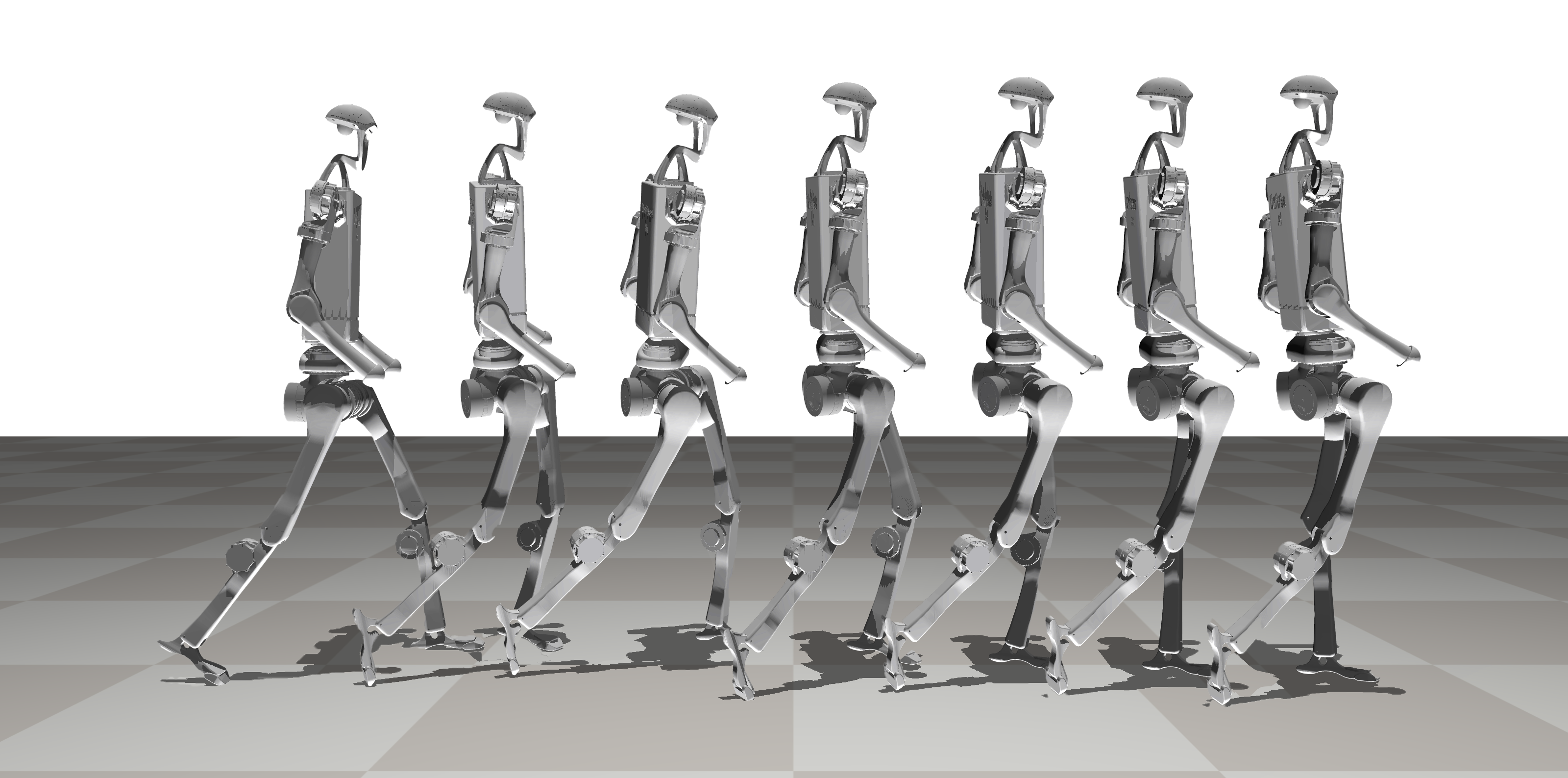}
        \caption{NIL trained on Kling v1.0}
        \label{fig:kling15}
    \end{subfigure}
    \caption{\textbf{Different versions of video diffusion models:} (a) NIL learns to walk using the newest Kling version for reference video generation; (b) Reference video generated by the older version of King results in walking with an unbalanced gait.}
    \label{fig:diffusion_models}
\end{figure*}

\begin{table}[!h]
\centering
\caption{\textbf{Video Diffusion Models}: We compare effectiveness of different video diffusion models in terms of generating reference videos.}
\label{table:diffusion_models}
\begin{tabular}{ccc} 
\toprule
\textbf{Model} & \begin{tabular}[c]{@{}c@{}}\textbf{Generated}\\\textbf{Frames}\end{tabular} & \begin{tabular}[c]{@{}c@{}}\textbf{NIL}\\\textbf{Performance}\end{tabular}  \\ 
\hline
Kling          & 150                                                                         & 396.1                                                                       \\
Pika v1.5      & 120                                                                         & 385.9                                                                       \\
Runway Gen-3   & 125                                                                         & 383.7                                                                       \\
OpenAI Sora    & 150                                                                         & 370.8                                                                       \\
SVD            & 24                                                                          & 366.5                                                                       \\ 
\hline
Expert         & -                                                                           & 400                                                                         \\
\bottomrule
\end{tabular}
\end{table}

Generated videos are provided in the Supplementary Materials. Quantitatively (see Table~\ref{table:diffusion_models}), Kling, despite exhibiting intermittent instabilities, yields the most visually plausible outputs and the highest NIL performance. Interestingly, even though Pika has shown limitations in physical plausibility \cite{bansal2024videophy}, it still leads to high imitation scores. We hypothesize that visual plausibility of the reference video is the most crucial property for NIL, as NIL is designed to refine physically implausible motions within the simulator.


\subsection{Improvements in Video Diffusion Models}
To examine the sensitivity of NIL to advancements in video diffusion models, we compare two versions of Kling: v1.0 and v1.6. Both versions are used to generate reference videos for the UnitreeH1 walking task. While the quantitative metrics are similar for both versions, qualitative results (see Figure~\ref{fig:diffusion_models}) reveal that the newer Kling v1.6 produces significantly more natural gaits. In contrast, the reference video from Kling v1.0 leads to an unbalanced gait with asymmetric leg movements.


This experiment underscores that even with small improvements in video diffusion models, the performance of NIL gets better. Therefore, better video diffusion models would enable NIL to learn more challenging tasks without using any collected/curated data.

\subsection{Continuous Control of Various Robots}
Finally, we test NIL on challenging locomotion tasks across multiple robotic embodiments: three humanoid platforms (Unitree H1, Talos, and Unitree G1) and a quadruped (Unitree A1). For each robot, NIL is trained using a single reference video generated by Kling AI (Pika for Unitree A1), and we compare its performance against upper baselines (AMP, DRAIL, GAIfO) as well as a lower baseline (BCO), all of which are trained with 25 curated expert demonstrations.

\begin{table}
\centering
\caption{\textbf{Robotic Control:} We evaluate NIL on challenging tobotic locomotion tasks across multiple robots.}
\label{table:robotic_control}
\begin{tabular}{ccccc} 
\toprule
                                                     & \begin{tabular}[c]{@{}c@{}}Unitree\\H1\end{tabular} & ~ ~Talos ~ ~ & \begin{tabular}[c]{@{}c@{}}Unitree\\G1\end{tabular} & \begin{tabular}[c]{@{}c@{}}Unitree\\A1\end{tabular}  \\ 
\hline
\begin{tabular}[c]{@{}c@{}}NIL \\(ours)\end{tabular} & \textbf{396.1}                                               & \textbf{352.8}~       & 356.9~                                              & \textbf{360.8}                                                \\ 
\hline
AMP                                                  & 393.5                                               & 231.1~       & \textbf{393.4 }                                              & 286.9                                                \\
DRAIL                                                & 11.7                                                & 1.9          & 7.3                                                 & 3.5                                                  \\
GAIfO                                                & 347.8                                               & 204.4        & 46.1                                                & 260.8                                                \\
BCO                                                  & 72.0                                                & 26.6         & 21.2~                                               & 30.3~                                                \\ 
\hdashline[1pt/1pt]
Expert                                               & 400~                                                & 400          & 400~                                                & 400~                                                 \\
\bottomrule
\end{tabular}
\end{table}

Table~\ref{table:robotic_control} presents quantitative results. For the Unitree H1 and Unitree A1, NIL not only achieves superior quantitative performance compared to AMP, it also produces a more natural and balanced walking gait. In contrast, for Unitree G1, even though NIL obtains competitive scores, AMP generates visually more natural locomotion. With the Talos platform, both NIL and AMP face significant challenges due to the robot’s complex morphology; however, NIL performs better and learns to move forward, albeit with less natural motion than desired.

\subsection{Summary of Experiments}
Overall, our experimental results confirm that NIL, by leveraging generated data and a discriminator-free imitation reward, effectively learns task-achieving policies across diverse robotic platforms. The ablation studies underscore the importance of the reward components, while the diffusion model comparison highlights that visually plausible generation, even if not physically perfect, is key for effective imitation. We also present that improvements in video diffusion models enables better performance of NIL. These findings demonstrate the potential of NIL as a promising alternative to conventional, data-intensive imitation learning approaches. Future improvements in video diffusion model could enable NIL to achieve more complex tasks, with different embodiments.
\section{Discussion and Future Directions}
\label{sec:discussion}
We introduce NIL as a first step towards eliminating the dependency on curated expert data in imitation learning. By leveraging video diffusion models to generate expert demonstrations on-the-fly, NIL not only removes the platform specific data collection but also achieves competitive performance across diverse robotic platforms. One of the key insights from our study is that the performance of NIL is closely tied to the quality of the generated videos. As improvements in video diffusion models continue to emerge, NIL naturally benefits from these advancements, leading to more realistic and robust behaviors. For example, our experiments with different versions of the Kling model clearly illustrate that even minor enhancements in video quality translate to significantly more natural gaits and smoother motion patterns.

Looking forward, several directions can further enhance the capabilities of NIL. First, integrating NIL as a pretraining step offers an exciting opportunity; the policies learned in a data-free manner can be fine-tuned using a small amount of curated data to boost motion naturalness, especially for complex morphologies where current methods face challenges. Also, extending NIL to more challenging tasks beyond locomotion is an exciting direction.

In summary, NIL’s performance is expected to improve in accordance with the rapid advancements in video diffusion technology. This synergistic relationship opens up a broader research horizon where data efficiency and generalization are paramount. We believe that our work lays a strong foundation for future research at the intersection of generative modeling and imitation learning, paving the way for increasingly sophisticated and autonomous robotic behaviors.
{
    \small
    \bibliographystyle{ieeenat_fullname}
    \bibliography{main}
}


\end{document}